# A method for nose-tip based 3D face registration using maximum intensity algorithm

Parama Bagchi , Debotosh Bhattacharjee, Mita Nasipuri, Dipak kr. Basu

Department of Computer Science and Engineering, MCKV Institute of Engineering, Kolkata-711204,
Department of Computer Science and Engineering, Jadavpur University, India, Kolkata-700032
paramabagchi@gmail.com, debotosh@ieee.org, mitanasipuri@gmail.com, dipakkbasu@gmail.com

*Abstract*— **In this paper we present a novel technique of registering 3D images across pose. In this context, we have taken into account the images which are aligned across X, Y and Z axes. We have first determined the angle across which the image is rotated with respect to X, Y and Z axes and then translation is performed on the images. After testing the proposed method on 472 images from the FRAV3D database, the method correctly registers 358 images thus giving a performance rate of 75.84%.**

## I. INTRODUCTION

Face detection and recognition has been the key areas of interest over the years . For a face which is oriented across any pose variation, in order to be correctly recognized, it should be perfectly registered . The goal of our proposed algorithm is to take an input 3D image across any pose orientations i.e. across X, Y and Z axes and return the final registered image. We have performed all the experiments on the FRAV3D database. The advantage of performing such experiments on 3D face database is that the 2D face images are very sensitive to lighting conditions or to pose and illumination changes. 3D faces overcome such shortcomings. 3D faces can even generate the complete texture information regarding a particular face in case of some scanners. With the rapid development of 3D technology, it is believed that 3D face recognition would be able to gain much popularity over 2D images [2]. The rest of the paper has been organized as follows. In Section II, we have discussed a comparison of our method over the already existing methods for 3D face registration. In Section III, we have discussed our proposed algorithm. In Section IV, a comparative analysis of our algorithm has been presented. Experimental results have been discussed in Section V, and finally, in Section VI, conclusions and future scope have been discussed.

## II. RELATED WORK

In this section, we will discuss some significant works done in the field of 3D face registration and thus compare our method with the existing 3D face registration technique. In [2], the authors have registered the face with the assumption that the nose tip has the maximum value across X axis and they have registered the face across X axis by fitting a plane across the nose-tip along X-axis. But, the nose-tip may not have the maximum value across X-axis. In [3], the authors have registered the images considering facial symmetry. Here also the authors have assumed, the maximum distance of the nose tip is from the curve ends and then registration is performed using the nose-tip. In [4], the authors have used an effective method to register the images depending upon their orientation but no pose variation across X, Y or Z axes is taken into account. In [5], the authors have used a joint shape and texture image to generate a set of region template detectors. The problem is that in their paper they have not dealt with pose variations across all possible orientations. In [6], the authors have detected face after registering by rotating the unregistered models by 45º. But the reason, how the angle is determined is not stated. If we go through existing popular registration techniques like ICP algorithm and Procrustes Analysis, then the difficulty with the methods are that, the unregistered models must be roughly registered initially, and then only ICP or Procrustes analysis could generate suitable results. Yet another problem with Procrustes Analysis is that, some predefined manual landmarks must be present for Procrustes Analysis to register two surfaces correctly. In FRAV3D database, there are no such manual landmarks available.

Here, in this paper we have calculated the rotational and translational parameters on the basis of which the registration process could be completed. It is actually worthwhile to mention here that the process of the registration adopted by us is a one-to all registration technique i.e we are registering the input image with all the images in frontal pose from the FRAV3D database and choosing the best image. It is to be mentioned that our technique implies a method of rough registration. In the following section we shall describe our proposed algorithm.

## III. PROPOSED TECHNIQUE

The present algorithm for registration consists of the following steps:-
- Facial Image Acquisition
- Thresholding

- Surface Smoothing
- Feature Localization
- Alignment and registration of models

A. *Facial Image Acquisition:-* The present technique uses FRAV3D database in face recognition. The Fig-1 below show samples of range images that we have taken for testing from the FRAV3D.

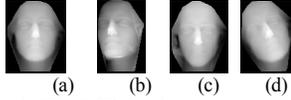

(a)   (b)   (c)   (d)

Fig1. Samples from the FRAV3D Database corresponding to a single person for frontal pose(a), image rotated about Y- axis(b), image rotated about X-axis(c), and image rotated about Z-axis(d).

B. *Preprocessing :-* First of all the 3D face of size[100 by 100] is cropped to size of dimensions[15 70]. Next, the image is thresholded by Otsu's thresholding algorithm After thresholding the 3D images obtained are shown as in Figure 2.

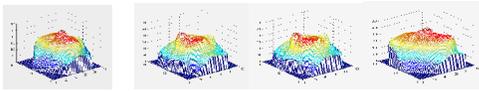

(a) Frontal pose (b) Rotated about y-axis(c) Rotated about x-axis (d) Rotated about z axis

Fig.2. Mesh-grids after thresholding

C. *Smoothing:-* In the next step the range image is smoothed by weighted median filter. After smoothing the 3D images obtained are as shown in Fig 3.

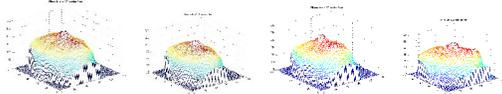

Fig 3. Mesh grids after smoothing

D. *Feature Localization*:- Faces have a common general form with prominent local structures such as eyes, nose, mouth, chin etc. Facial feature localization is one of the most important tasks of any facial classification system.

A. *Surface Generation:-*The next part of the present technique concentrates on generating the surface of this 3D mesh image. For the nose tip localization we have used the maximum intensity concept as the tool for the selection process. Each of the 472 faces (including rotation in any direction in 3D space namely about x-axis, y-axis and z-axis) in the FRAV3D database has been next inspected for localizing the nose tip. A set of fiducial points are extracted from both frontal and various poses of face images using a maximum intensity algorithm. As shown in Fig.5, the nose tips have been labeled on the facial surface, and accordingly, the local regions are constructed based on these points. The maximum intensity algorithm used for our purpose is as given below:-
 Function Find_Maximum_Intensity (Image)
  Step1:- Set max to 0
  Step 2:- Run loop for I from 1 to width (Image)
  Step 3:- Run loop for J from 1 to height (Image)
  Step 4:-  Set val to sum(image(I-1:I+1,J-1:J+1))
  Step 5:-  Check if val is greater than max.
  Step 6:-  Set val to val2

  Step 7: - End if
  Step 8:- End loop for I
  Step 9:- End loop for J

The 3D surfaces thus generated for FRAV3D database are shown in Fig 4.

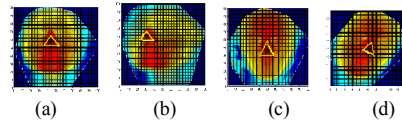

(a)   (b)   (c)   (d)

Fig 4. Surface Generation with nose localized from the FRAV3D Database.

E. *Alignment and Registration of Models:-* After feature localization, based on the extracted features we have to align the extracted face model. This alignment with the major axes greatly simplifies the task for registration. 3D registration comprises of two major steps :-

1) *Rotation*
2) *Translation*

1) Rotation:- This step essentially finds out the angle by which the model to be registered is to be rotated. Since our technique is nose-tip based rotation so we find out a method to rotate our unregistered model by finding out the angle of rotation. Any input model can be registered against 3 axes :- X,Y and Z. So first, we shall consider rotation against Z axes. As we have implied that our method is a one-to-all registration technique, so we have basically registered our image model with the neutral models corresponding to a particular individual from the FRAV3D database. So, at first, let us consider the image model rotated about Z axis.

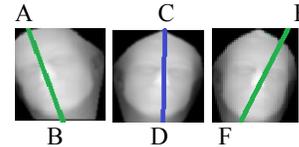

A     C     E

B     D     F

Fig 5. Image rotated with respect to left orientation of Z-Axis, Image in frontal pose and image rotated with respect to right orientation of Z-axis.

The process of registration could be simplified to finding the angle between the AB and CD or CD and EF . We have plotted the coordinates of the nose-tips in case of rotated(R) and frontal images(F). The result of the plot is as shown in Fig 6.

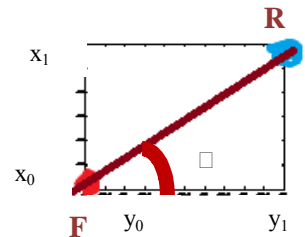

Fig 6. Points plotted in case of frontal and rotated pose.

Now our task of finding out the angle becomes simpler. We just have to find out □. So we apply the formula:-

$\tan\theta = (+/-)\dfrac{x_1 - x_o}{y_1 - y_o}$. Here, the notations $x_1, y_1, x_o, y_o$ denotes the displacements in x directions and y directions of the nose-tips of rotated pose and frontal poses respectively. If the image is rotated with respect to left-orientation, then the resulting image is to be rotated by a negative angle else if, in case, the image is rotated with respect to right-orientation, then the resulting image is to be rotated by a positive angle. Then, the un-registered image is rotated by the angle □ and the rotation matrix is as follows:-

$S'_i = M_z * S_i$ where

$$M_z = \begin{bmatrix} \cos(theta) & -\sin(theta) & 0 & 0 \\ \sin(theta) & \cos(theta) & 0 & 0 \\ 0 & 0 & 1 & 0 \\ 0 & 0 & 0 & 1 \end{bmatrix}$$

and $S_i$ is the original pointcloud. Since the feature that we have selected is the nose-tip, so based on the nose-tip we have performed a coarse alignment of the models.

   2) Translation: - Finally, the data points are translated to the origin so that the final analysis of the registered image could be done accurately. After rotation and translation, the resulting image is shown in Fig 7.

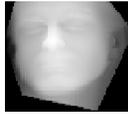

Fig 7. Registered Image after rotation with respect to the z axis.

The algorithm runs till the eyes of the rotated image are in the same horizontal line which proves that the image has been perfectly registered.

Next, let us consider the situation where the image is rotated with respect to the Y-axis.

  1) Rotation:- Let us consider the following image rotated about Y-axes.

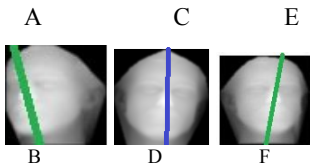

Fig 8 . Image rotated with respect to left orientation of y-Axis, Image in frontal pose and image rotated with respect to right orientation of y-axis.

 Here, we have to find the angle between the lines AB and CD or CD and EF. In this case, we have to find the angle between the points the lines passing through the point R which is the nose-tip of the rotated pose and the line passing through F which is the nose-tip of the frontal pose. Since here the images are rotated with respect to y axis , so the y coordinates will vanish. The result of the plot is as shown in Fig 9.

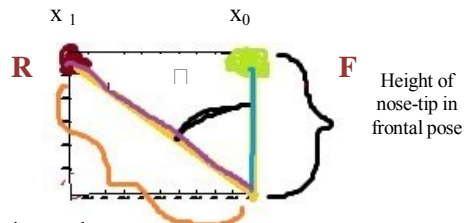

Fig 9. Points plotted in case of frontal and rotated pose.

Now , we  just have to find out □ . So we apply the formula:-

$\tan \theta = (+/-) \dfrac{x_1 - x_0}{height}$ .Here, the notations $x_1$, $x_o$ denotes the displacements in y directions and height is the height of the nose tip in frontal pose. If the image is rotated with respect to left- orientation, then the resulting image is to be rotated by a negative angle else if, in case, the image is rotated with respect to right- orientation, then the resulting image is to be rotated by a positive angle. To eliminate the tilt along Y-axis, the image has to be rotated and that rotation can be obtained by multiplying the original pointcloud image by the matrix given as follows:-

$S'_i = M_y * S_i$ where

$$M_y = \begin{bmatrix} -\sin(theta) & 0 & \cos(theta) & 0 \\ 0 & 1 & 0 & 0 \\ -\sin(theta) & 0 & \cos(theta) & 0 \\ 0 & 0 & 0 & 1 \end{bmatrix}$$

   2) Translation :- Finally , the data points are translated to the origin so that the final analysis of the registered image could be done accurately. After rotation and translation the resulting image is shown in Fig 10.

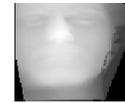

Fig 10. Registered Image after rotation with respect to the y axis.

The image is rotated till the x coordinate of the rotated image is quite equal to the x coordinate of the nose tip of the frontal image against which the image has been registered.

We now consider the situation where the image is rotated with respect to the X-axis as shown in Fig 11.

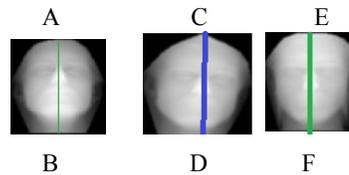

Fig 11 . Image rotated with respect to left orientation of x-Axis, Image in frontal pose and image rotated with respect to right orientation of y-axis.

 1)Rotation:- Here, in this case the process of registration was found by  finding the angle between the lines AB and CD or CD and EF. We have plotted the coordinates of the nose-tips in case of rotated denoted by R and frontal images denoted by F. Since here the images are rotated with respect to x axis , so the x coordinates will vanish. The result of the plot is as shown in Fig 12.

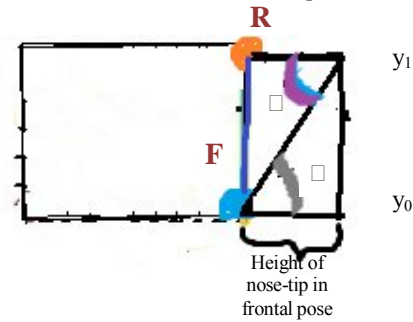

Fig 12.Points plotted in case of frontal and rotated pose

The case where the image is rotated with respect to x axis is a complicated one. We need to find the angle marked □, which according to the theory of parallel line is equal to the angle marked θ. Next, we should keep in mind that we are registering against all frontal images, so angle should be found out against all frontal images. We just have to find out θ. So we apply the formula:-

$\tan\theta = (+/-) \dfrac{y_1 - y_0}{depth}$ .Here, the notations $y_1$, $y_o$ denotes the displacements in y directions and depth is the height of the nose tip in frontal pose. If the image is rotated with respect to left- orientation, then the resulting image is to be rotated by a negative angle else if, in case, the image is rotated with respect to right- orientation, then the resulting image is to be rotated by a positive angle. To eliminate the tilt along Y-axis, the image has to be rotated and that rotation can be obtained by multiplying the original pointcloud image by the matrix given as follows:-

$S_i' = M_x * S_i$ where

$M_x = \begin{pmatrix} 1 & 0 & 0 & 0 \\ 0 & \cos(theta) & -\sin(theta) & 0 \\ 0 & \sin(theta) & \cos(theta) & 0 \\ 0 & 0 & 0 & 1 \end{pmatrix}$

2) Translation :- The data points are translated to the origin so that the final analysis of the registered image could be done accurately. After rotation and translation the resulting image is shown in Fig 13.

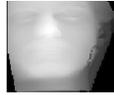

Fig 13. Registered Image after rotation with respect to the x axis.

The image is rotated till the x coordinate of the rotated image is quite equal to the x coordinate of the nose tip of the frontal image against which the image has been registered. The algorithm which we have implemented for registration for of the 3D face are as given below:-

**Algorithm 1**

Step 1:- Input the 3D face across any pose
Step 2:- Extract the nose-tip using maximum- intensity algorithm
Step 3:- Find out the angle necessary for registration
Step 4: - Rotate and translate the resulting point-cloud
Step 5:- Now extract the eyes for frontal and rotated image by curvature analysis[6]
Step 6:- For images rotated about Z axes check if the eye-corners are in the same horizontal line or not i.e. if the difference in x coordinates of eye-corners for images rotated about z axes <2 , then the image is perfectly registered.
Step 7:-For images rotated about x-axis, extract the nose-tips from both frontal and rotated poses. Now check if the x-coordinate of the nose-tip for the rotated image- xcoordinate of the nose-tip for frontal image <2 then the image is perfectly registered.
Step 8:- For images rotated about y-axis, extract the nose-tips from both frontal and rotated poses. Now check if the x-coordinate of the nose-tip for the rotated image- xcoordinate of the nose-tip for frontal image <2 then the image is perfectly registered

**End of Algorithm**

IV. COMPARATIVE ANALYSIS

Here we present a comparative analysis of how our method of registration outperforms the method specified in [7]. The authors in [7] have also worked on FRAV3D database. The comparative analysis between the technique in[7] and the present proposed technique has been enlisted in the Table I. From the table I, it is obvious that our technique outperforms the technique mentioned in [7] in terms of complexity analysis, response time and performance rate.

TABLE I. COMPARATIVE STUDY BETWEEN THE PROPOSED TECHNIQUE AND THE TECHNIQUE MENTIONED IN[7]

| Table 1 | Our proposed algorithm | Method for registration in [7] |
|---|---|---|
| Response time for registration | Max time required:- 11.8sec Min time required:-8 sec | Not specified |
| Complexity analysis of algorithm | $O(n)+O(n^2)+O(n^2)$ if n is the number of 3D points | Not specified |
| Landmark used in registration | Only nose-tip is used for finding of angle of rotation | Both eyes and nose-tip used |
| Performance rate | 75.84% performance rate of registration | A proposed rate of 80% is given for recognition only. |

V. EXPERIMENTAL RESULTS

The Fig-14 shows some of our registration results after applying the proposed algorithm.

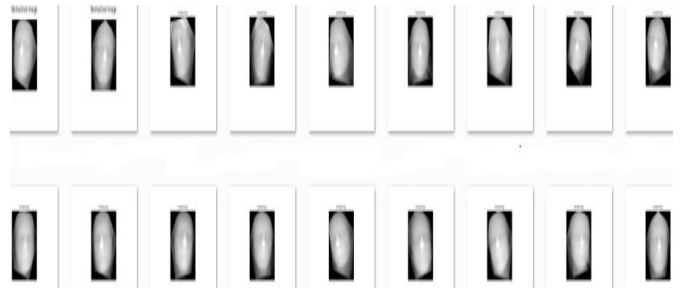

Fig 14. Registered image gallery from the FRAV3D after applying the proposed method

After applying our method for registration on images from the FRAV3D database we obtain results as in Tables II,III and IV :-

TABLE II. REGISTRATION RESULTS ACROSS Z AXES

| Table 2 | Registration Results across Z axes | | | | |
|---|---|---|---|---|---|
| | Angles of Rotation | No of faces rotated across Z axes | No. of faces correctly registered | % of Success | % of Failures |
| 1. | +18 | 20 | 15 | 67.7% | 32.22% |
| 2 | -18 | 20 | 12 | | |
| 3 | +30 | 10 | 10 | | |
| 4 | -30 | 20 | 15 | | |
| 5 | +38 | 10 | 8 | | |
| 6 | -38 | 10 | 10 | | |
| 7 | +40 | 9 | 5 | | |
| 8 | -40 | 9 | 5 | | |

TABLE III REGISTRATION RESULTS ACROSS Y AXES

| Table 3 | Registration Results across Y axes | | | | |
|---|---|---|---|---|---|
| | Angles of Rotation | No of faces rotated across Y axes | No. of faces correctly registered | % of Success | % of Failures |
| 1. | +5 | 50 | 40 | 79.66% | 20.33% |
| 2 | -5 | 50 | 40 | | |
| 3 | +18 | 50 | 40 | | |
| 4 | -18 | 50 | 40 | | |
| 5 | +40 | 18 | 14 | | |
| 6 | -40 | 18 | 14 | | |

TABLE IV . REGISTRATION RESULTS ACROSS X AXES

| Table 4 | Registration Results across x axes | | | | |
|---|---|---|---|---|---|
| | Angles of Rotation | No of faces rotated across Y axes | No. of faces correctly registered | % of Success | % of Failures |
| 1. | +30 | 20 | 10 | 79.66% | 20.33% |
| 2 | -30 | 20 | 20 | | |
| 3 | +38 | 20 | 10 | | |
| 4 | -38 | 20 | 10 | | |
| 5 | +40 | 20 | 20 | | |
| 6 | -40 | 28 | 20 | | |

## VI. CONCLUSION AND FUTURE SCOPE

This paper demonstrated that it is possible to calculate the translational and rotational parameters necessary for registration of 3D face images. Although the results obtained from the registration are an average one, but still the approach can prove to be useful in case of other 3D face databases also.

However, the present work has some limitations. We have to find out some robust methodologies for discarding outliers in case of very noisy images. Also, as a part of our future work, we shall develop a more robust method of registration. In future, we plan to test the result of our registration with the help of some standard classifiers like PCA,LDA and SVM.